\documentclass[11pt]{article}

\usepackage[final]{acl}
\usepackage{xspace}
\usepackage{times}
\usepackage{latexsym}
\newcommand{\qwq}{{QwQ-32B}\xspace}
\newcommand{\distill}{{Distill-14B}\xspace}
\newcommand{\qwen}{{Qwen3-32B}\xspace}
\newcommand{\gsmdata}{{GSM8K}\xspace}
\newcommand{\mathdata}{{MATH500}\xspace}
\newcommand{\aimedata}{{AIME2024}\xspace}
\usepackage{tabularx}
\usepackage[T1]{fontenc}
\usepackage{newfloat}
\usepackage{listings}
\usepackage{booktabs}
\usepackage{multirow}
\usepackage{amsmath}
\usepackage[utf8]{inputenc}
\usepackage{algorithm}
\usepackage{algorithmic}
\usepackage{soul}

\usepackage{microtype}

\usepackage{inconsolata}

\usepackage{graphicx}

\title{When Can Large Reasoning Models Save Thinking? \\ Mechanistic Analysis of Behavioral Divergence in Reasoning}

\author{
    \textbf{
    Rongzhi Zhu\textsuperscript{1},
    Yi Liu\textsuperscript{1},
    Jiancheng Wang\textsuperscript{2}\thanks{Corresponding author.},
    Xiangyu Liu\textsuperscript{1},
    Zequn Sun\textsuperscript{1}\footnotemark[1]
    } \\
    \textbf{
    Yiwei Wang\textsuperscript{3},
    Yu Deng\textsuperscript{2},
    Zijian Zhou\textsuperscript{2},
    Wei Hu\textsuperscript{1,4}
    } \\
    \textsuperscript{1}State Key Laboratory for Novel Software Technology, Nanjing University, China  \\
    \textsuperscript{2}China Unicom Software Research Institute, China\\
    \textsuperscript{3}University of California, Merced, USA \\
    \textsuperscript{4}National Institute of Healthcare Data Science, Nanjing University, China \\
    \texttt{\{rzzhu.nju, yiliu07.nju, xyl.nju, wangyw.evan\}@gmail.com} \\
    \texttt{\{sunzq, whu\}@nju.edu.cn} \\
    \texttt{\{dengy63, wangjc161, zhouzj121\}@chinaunicom.cn}
}

\begin{document}
\maketitle
\begin{abstract}
Large reasoning models (LRMs) have achieved remarkable success on complex tasks, yet their tendency to ``overthink'' leads to inefficiencies.
Although ``save-thinking'' prompts are intended to mitigate this issue, we find that LRMs still frequently enter the ``Still-thinking'' mode instead of the expected ``No-thinking'' mode, especially on difficult queries.
To analyze this behavioral divergence, we examine LRMs from three perspectives: confidence at the thinking-termination boundary, divergence in internal attention distributions, and attention allocation across prompt segments.
We find that high perplexity is associated with later Still-thinking behavior, and that Still-thinking cases allocate more attention to the original question.
Based on these observations, we propose an attention intervention method to regulate this behavior.
While this intervention suppresses explicit thinking, it also causes a drop in accuracy, suggesting that the suppressed reasoning behavior is often useful for correctness.
Our work provides confidence- and attention-level evidence for this behavior, highlighting the trade-off between instruction following, inference efficiency, and reasoning correctness.
\end{abstract}

\section{Introduction}\label{sect:intro}

Large language models (LLMs) have demonstrated remarkable performance across a wide range of domains \cite{GPT4,Qwen25,deepseekv3}.
When integrated with retrieval-augmented generation \cite{RAG,chainrag}, supervised fine-tuning (SFT) \cite{finetune,lora}, or knowledge editing \cite{alphaedit}, LLMs achieve improved capabilities in specialized domains beyond their original pretraining scope.
However, LLMs may still struggle with complex reasoning tasks \cite{llm_math}. 
The breakthrough of large reasoning models (LRMs) like OpenAI-o1 \cite{OpenAI2025o1} highlights the benefits of scaling test time,
enabling extended reasoning and the generation of comprehensive thoughts. 
This approach can significantly improve the accuracy on complex reasoning tasks.
Recently, an increasing number of LRMs, such as QwQ \cite{QwQ} and DeepSeek-R1 \cite{r1}, have been developed.
These LRMs can be categorized based on their training paradigms: 
one class leverages SFT on chain-of-thought (CoT) data \cite{s1}, 
while the other is trained directly through reinforcement learning (RL). 
LRMs demonstrate superior performance compared to LLMs on various reasoning tasks \cite{survey_long}.

\begin{quote} 
    // \textit{instruction segment}\newline
    \textbf{User}: Your final answer should follow immediately after the phrase ``The final answer is''.  \newline
    // \textit{question segment}  \newline
    \textbf{Question}: [{Question}].  \newline
    // \textit{thought segment} \newline
    \textbf{Assistant}: \newline 
    \texttt{<think>} \newline
    Okay, I think I have finished thinking. \newline
    \texttt{</think>}
\end{quote}

Despite the significant improvement in reasoning capabilities, LRMs usually suffer from the overthinking problem. 
They tend to generate excessively long reasoning chains, even for simple tasks \cite{2+3},
which leads to computational inefficiency and sometimes worse accuracy \cite{StopOverthinkingSurvey}.
Most LRMs, such as \qwq{}, generate tags like \texttt{<think>} and \texttt{</think>} to encapsulate the thinking process, followed by the final answer.
To address the overthinking issue, \citet{skip_reason} propose prompting LRMs to bypass thinking by pre-filling the segment between \texttt{<think>} and \texttt{</think>}.
The prompt template for saving thinking consists of the instruction segment, question segment and thought segment as shown below:

\begin{figure}[t]
  \includegraphics[width=\linewidth]{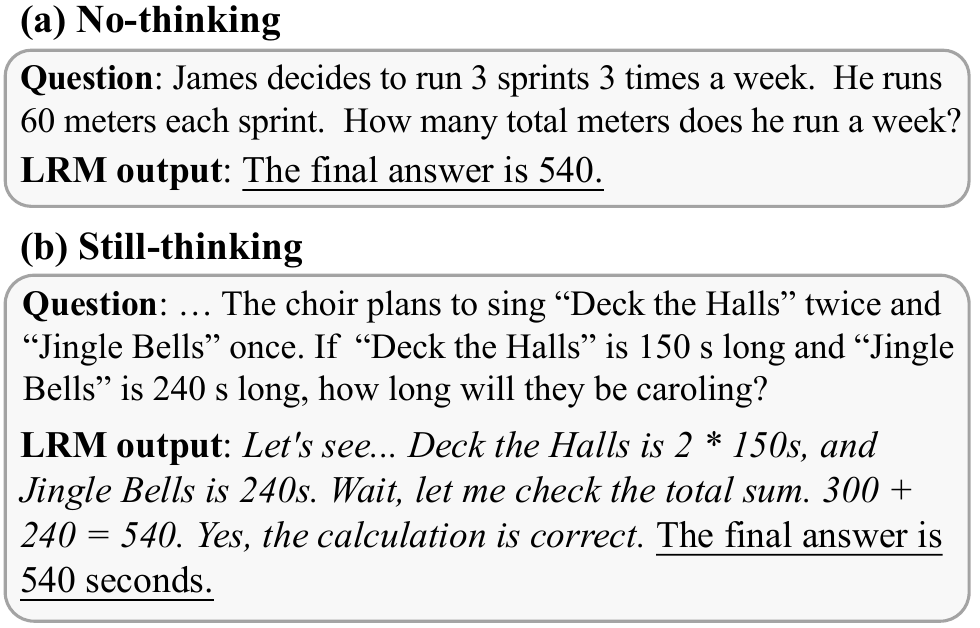}
  \caption{Examples of the NT and Thinking modes of LRMs under the save-thinking instructions. The final answers are underlined. The thought is marked in italics.}
  \label{fig:example}
\end{figure}

However, our empirical findings show that when this prompt is applied to LRMs, their behavior is inconsistent: 
while they occasionally bypass the thinking process as intended, they often reinitiate thinking unexpectedly.
This inconsistency raises a key question: \textit{what internal signals are associated with LRMs overriding the ``save-thinking'' instruction, and how are these signals related to task difficulty?}
To investigate this problem, we categorize the behaviors of LRMs under the ``save-thinking'' instruction into two modes, i.e., No-thinking (NT) and Still-thinking (ST), based on whether the model exhibits clear reasoning behaviors, such as backtracking or verification. 
Specifically, we check for keywords and phrases such as ``let me check'', ``wait'' and others in the response to make this distinction:

\begin{itemize}
    \item \textbf{No-thinking}: The LRM bypasses further thinking and directly generates the answer.
    
    \item \textbf{Still-thinking}: The LRM overrides the instruction and re-engages in a thinking process, as evidenced by explicit reasoning indicators.
\end{itemize}

To explore the distinctions between two modes, we examine LRMs from three perspectives: 
confidence in thinking termination, 
attention divergence during generation, 
and attentional focus on prompt segments.
We further analyze how these modes affect performance and output length.
Our contributions and findings are summarized as follows:
 
\begin{itemize}
    \item \textbf{Difficulty-related behavioral divergence:}
    We identify that the failure to ``save thinking'' is not random.
    Instead, the Still-thinking mode becomes more frequent as task difficulty increases, suggesting that LRMs are more likely to override the instruction on difficult queries.
    
    \item \textbf{Confidence- and attention-level analyses:}
    We analyze this divergence from three perspectives across different LRMs.
    First, high perplexity at the termination boundary is strongly associated with later Still-thinking behavior.
    Second, attention distributions for NT and ST cases become separable in early layers.
    Third, ST cases allocate more attention to the original question and less to the pre-filled thought segment, which we describe as retrospective attention allocation.
    
    \item \textbf{Intervention and performance analysis:}
    We introduce an attention intervention that limits attention to the question segment during generation.
    This intervention can suppress explicit thinking in some cases, but it also reduces accuracy on difficult samples.
    These results suggest that the suppressed reasoning behavior is often useful for correctness, highlighting the trade-off between instruction following, inference efficiency, and reasoning accuracy.
\end{itemize}

Our code, results and the annotations of the two thinking modes in the \gsmdata{}, \mathdata{} and \aimedata{} datasets will be publicly available to support future research.

\section{Related Work}

\subsection{Large Reasoning Models}

LLMs have demonstrated stronger capabilities with larger scale \cite{llms}, yet they still struggle with complex reasoning tasks like mathematics and code generation \cite{llm_math,rl_math,mathsurvey}. Recent work has found that instead of continuously scaling model size and training data, scaling the model's thinking time and having it generate chain-of-thought reasoning—similar to human thinking—can significantly improve the accuracy of complex tasks \cite{scaling}. Consequently, LRMs such as OpenAI-o1 \cite{OpenAI2025o1}, DeepSeek-R1 \cite{r1}, and QwQ \cite{QwQ} have emerged, specifically designed to produce structured reasoning processes before the final answer. These LRMs improve performance on challenging reasoning benchmarks by generating longer thinking processes that involve considering multiple potential solutions and backtracking from errors \cite{survey_long}. 

\begin{table}[t]
\centering
\resizebox{\linewidth}{!}{\Large
\begin{tabular}{l cc cc cc}
\toprule
\multirow{2}{*}{Dataset} & \multicolumn{2}{c}{QwQ-32B} & \multicolumn{2}{c}{Qwen3-32B} & \multicolumn{2}{c}{\distill} \\
\cmidrule(lr){2-3} \cmidrule(lr){4-5} \cmidrule(lr){6-7}
& \multicolumn{1}{c}{NT} & \multicolumn{1}{c}{ST} & \multicolumn{1}{c}{NT} & \multicolumn{1}{c}{ST} & \multicolumn{1}{c}{NT} & \multicolumn{1}{c}{ST} \\
\midrule
\gsmdata{}  & 970 & 349 & 1289 & 30 & 1245 & 74 \\
\mathdata{} & 118 & 382 & 416 & 84 & 246 & 254 \\
\aimedata{} & 2 & 28 & 12 & 18 & 4 & 26 \\
\bottomrule
\end{tabular}}
\caption{Data statistics of test samples grouped by the thinking modes across LRMs and datasets.}
\label{tab:dataset-dist}
\end{table}

\subsection{Mechanistic Interpretability of LRMs}

As the capabilities of LRMs increase, understanding the mechanisms of their internal reasoning processes becomes crucial. Recent research applies methods from mechanistic interpretability to investigate the specific functions of internal components. One line of research investigates the LRM's confidence during reasoning. For instance, \cite{early_exit} calculates the confidence score when generating special tokens and allows for an early exit if the confidence exceeds a certain threshold. Similarly, \cite{confidence_hard} found a strong correlation between a LRM's confidence and its reasoning accuracy. Some other studies focus on the principles behind the emergence of reasoning abilities. \cite{rast} discovered that the decoding paths of LRMs and base models are largely identical, diverging only when decoding special tokens that signal additional reasoning steps, while \cite{entropy} found that strong reasoning capabilities can be achieved by minimizing the entropy of the base LRM's reasoning process during training. Another line of work focuses on the analysis of specific internal components. For example, \cite{who_reason} identified the attention projector layer as the primary component responsible for reasoning, and \cite{internal_bias} pointed out that biases in the LRM's internal knowledge can lead to the phenomenon of ``overthinking''. Additionally, \cite{hidden} noted that LRMs may produce reasoning of varying lengths for the same problem.
These long and short reasoning paths correspond to distinct clusters in the hidden state space.

\section{Mechanistic Analysis of Thinking Modes}
\label{sec:3}
In this section, we investigate internal signals associated with LRMs overriding ``save-thinking'' instructions.
We first introduce the experimental setup used in our analysis.
Then, we present confidence- and attention-level analyses from three aspects.
Finally, we introduce an intervention method to examine the role of attention allocation.

\paragraph{LRMs selection.}
For a comprehensive analysis, we select three open-source LRMs with different training paradigms:
\qwq{} and \qwen{}, both trained via RL, 
and DeepSeek-R1-Distilled-Qwen-14B (hereafter \distill), which is trained using SFT on distilled data. 
Our findings indicate that the internal states of LRMs also differ depending on their training methodology.

\paragraph{Datasets and annotations.}
Our experiments and analysis are conducted on three mathematical reasoning datasets with varying levels of difficulty: the test sets of GSM8K, MATH500, and AIME2024.
All three datasets require multi-step reasoning.
Specifically, for each question, we employ the ``save-thinking'' prompt detailed in Section 1 to direct the models to provide an immediate response.
We use GPT-4.1\cite{gpt4.1} for an initial classification of the outputs into NT and ST modes, followed by manual checks and corrections on ambiguous cases.
To assess annotation reliability, we further compare our final labels with independent human annotations and Qwen3-Max\cite{qwen3} annotations on Qwen3-32B outputs from MATH500.
The pairwise agreement rates are high: 98.2\% for final labels versus human annotations, 96.4\% for final labels versus Qwen3-Max, and 95.4\% for human annotations versus Qwen3-Max.
The statistics of the labeled data are presented in Table~\ref{tab:dataset-dist}.

\paragraph{Behavioral divergence and difficulty perception.}
As shown in Table~\ref{tab:dataset-dist}, the distribution of thinking modes reveals a critical insight: the tendency to override the ``save-thinking'' instruction is not random, but strongly correlated with task difficulty.
Across all models, the proportion of ``Thinking'' responses rises sharply as the dataset difficulty increases from \gsmdata{} to \aimedata{}.
For instance, \qwen{} follows the NT instruction in 97.7\% of cases on the simpler \gsmdata{}, but this compliance drops to only 40\% on the highly complex \aimedata{}.
Furthermore, \qwq is far more prone to re-engaging in thinking, doing so in over $28\%$ of cases.
This behavioral divergence, which is clearly influenced by both the model architecture and training, motivates our analysis of their internal states to understand the underlying mechanisms.

\begin{figure*}[t]
\centering
\includegraphics[width=\textwidth]{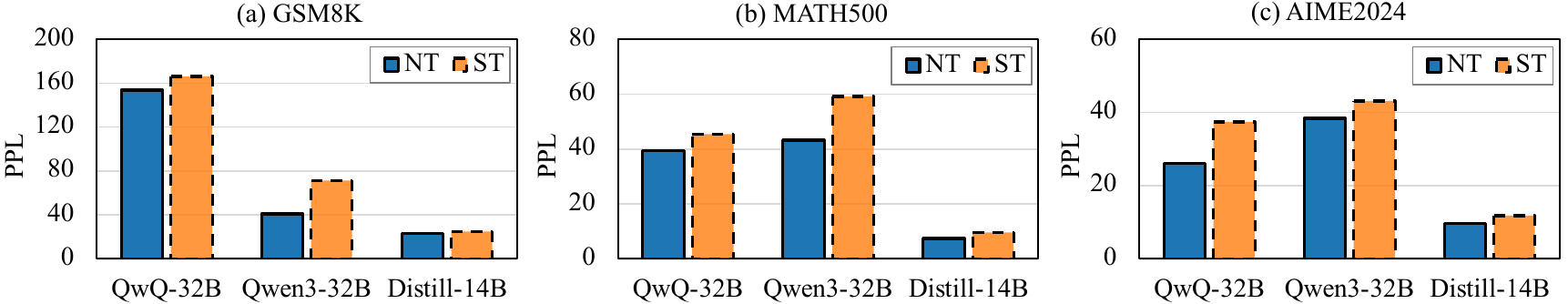}
\caption{PPL results across datasets and thinking modes. Lower values indicate higher confidence.}
\label{fig:complexity_results}
\end{figure*}



\subsection{Confidence in Thinking Termination}

\paragraph{Motivation and methodology.}
Our prompt includes a pre-filled thinking segment: ``\texttt{<think>} Okay, I think I have finished thinking. \texttt{</think>}'' to guide LRMs toward immediate answer generation.
The \texttt{</think>} token marks the boundary between reasoning and answering.
We posit that the predictive state at this juncture reflects the certainty about resolving the question without further reasoning. 
Therefore, a high uncertainty at this boundary likely serves as an internal signal that precedes the decision to override the instruction.

Specifically, we perform a forward pass on the pre-filled thinking segment to quantify the LRM's certainty when it is instructed to terminate its reasoning process. 
Instead of analyzing a single distribution, we measure the performance on the ground-truth tokens of the pre-filled phrase itself (i.e., ``Okay, I think I have finished thinking.''). 
To this end, we employ \textit{Perplexity} (PPL) \cite{PPL} as the evaluation metric,
which reflects how well the model predicts this given sequence. 
A higher perplexity indicates that the model is more ``surprised'' by the sequence, reflecting greater internal uncertainty or hesitation to conclude its thinking.

The Perplexity of a ground-truth subsequence $Y_{s:e} = (y_s, \dots, y_e)$ is calculated from its conditional probabilities $p_i = p(y_i|y_{<i})$. 
To ensure numerical stability, probabilities below a threshold $\epsilon$ (e.g., $10^{-9}$) are filtered out. 
Let $\mathcal{P}'_{s:e}$ be the set of these filtered probabilities:
\begin{equation}
\label{eq:filtered_probs}
\mathcal{P}'_{s:e} = \{ p_i \mid i \in [s, e] \text{ and } p_i > \epsilon \}.
\end{equation}
The PPL is then computed as the reciprocal of the geometric mean of these probabilities, implemented in log-space for stability. Let $N' = |\mathcal{P}'_{s:e}|$, the perplexity is given by:
\begin{equation}
\label{eq:perplexity}
\text{PPL}(Y_{s:e}) = \exp\big( -\frac{1}{N'} \sum_{p \in \mathcal{P}'_{s:e}} \ln(p) \big).
\end{equation}

\paragraph{Results and analysis.}
The results, illustrated in Figure~\ref{fig:complexity_results}, reveal a clear and consistent pattern across all three LRMs. 
For questions where the model successfully adheres to the NT mode, the PPL is significantly lower, indicating high confidence in terminating the thought process. 
In contrast, for questions that trigger the ST mode, the models exhibit sharply higher PPL.
On average, the PPL in the ST mode exceeds that in the NT mode by 41.72\% for \qwen{}, 22.15\% for \qwq{} and 19.25\% for \distill{}.
We further conduct two-tailed independent $t$-tests on GSM8K and MATH500.
As reported in Appendix~\ref{app:ppl-significance}, the NT--ST PPL difference is statistically significant in five out of six settings and significant at $p < 0.01$ for all three models on MATH500.
This supports the observation that high perplexity at the termination boundary is associated with later ST behavior.
For more difficult samples, the model tends to show higher PPL at this boundary, which correlates with the subsequent initiation of the ST mode.

Beyond the relative gap, we observe a distinct difference in absolute PPL values driven by training paradigms. 
\distill{}(SFT-based) demonstrates much lower overall PPL, indicating high confidence.
We hypothesize that SFT encourages the imitation of single correct paths, leading to decisive but potentially overconfident predictions.
In contrast, the RL-trained models (\qwq and \qwen{}) exhibit significantly higher baseline perplexity.
This suggests that RL training fosters a more ``reflective'' internal state. 

\paragraph{Findings.}
Our analysis reveals two key insights. 
First, the PPL gap serves as a reliable signal for reasoning demand: 
low perplexity indicates readiness to answer, while high perplexity indicates hesitation. 
Second, this uncertainty mechanism is shaped by training: RL training preserves a higher degree of uncertainty compared to the lower-entropy state induced by SFT.

\subsection{Attention Distribution Divergence}

\paragraph{Motivation and methodology.}

Since the model's confidence varies significantly between modes, we next investigate whether its internal attention states also diverge.
Specifically, does the model process ``hard'' (ST-triggering) queries differently from ``easy'' (NT-compliant) ones, and when does this divergence occur?
To quantify this, we measure the separation between the attention distributions of the NT and ST response groups using the \textbf{Davies-Bouldin (DB) Index} \cite{db_index} for each layer.
A lower DB Index indicates clearer, more distinct clustering, implying a sharper separation between the two modes.
To ensure a robust measurement across varying sample sizes, we pool samples from all three datasets (\gsmdata{}, \mathdata{}, and \aimedata{}) to calculate the global separation index for each LRM.

\begin{figure}[th] 
  \centering 
  \includegraphics[width=\linewidth]{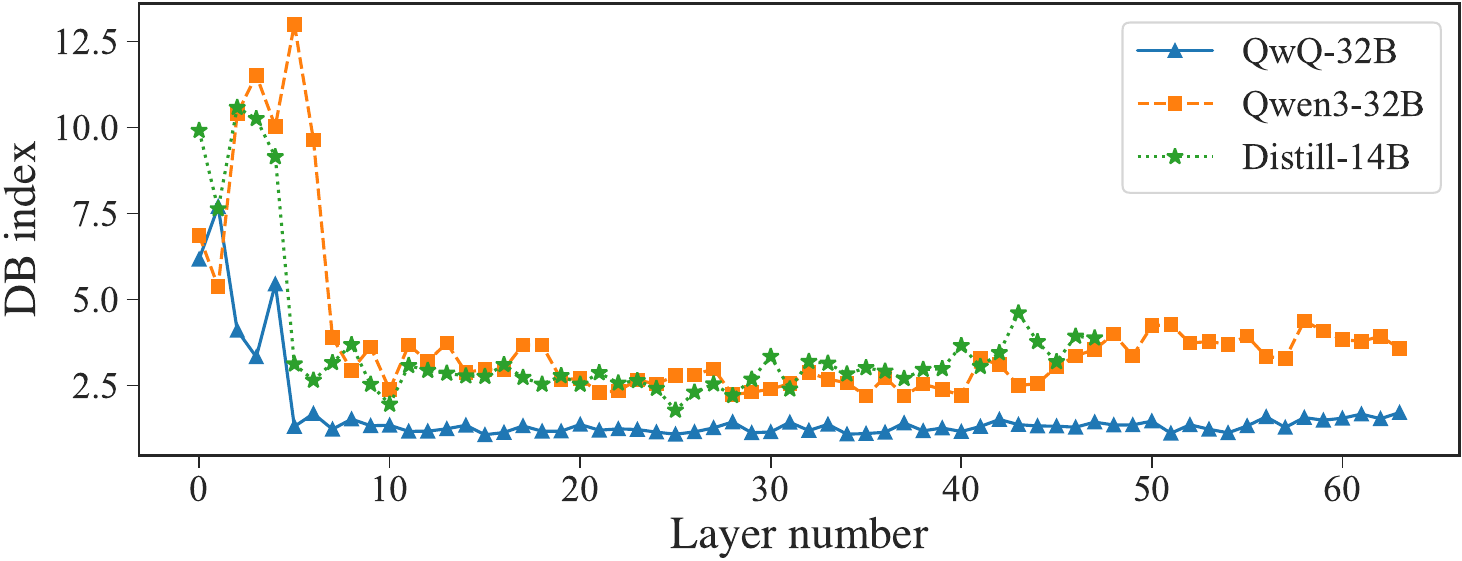} 
  \caption{DB Index calculated between NT and ST modes for each model. The index is computed layer-wise on a combined pool of all three datasets. Lower index is better. } 
  \label{fig:db_index} 
\end{figure}

\paragraph{Results and analysis.}
Figure~\ref{fig:db_index} illustrates the layer-wise DB Index trends.
We observe a sharp drop in the DB Index within the first 5--10 layers across all models, followed by a sustained low plateau.
This trajectory reveals that the divergence between ST and NT modes is an early-stage phenomenon.
The LRM performs an early assessment of the input query's complexity, establishing divergent processing pathways well before the actual generation begins.

While the overall trend is consistent, the three LRMs exhibit distinct characteristics. 
\qwq{} consistently achieves the lowest DB Index across the middle layers, suggesting it maintains the most clearly separable internal states for the two thinking modes. 
Compared to \qwq{}, which was also trained with RL, \qwen{} generally shows the highest DB Index, indicating that its internal states for NT and ST modes are less distinct.
We hypothesize that this reduced separation may stem from its training process,
which includes adaptive thinking strategies and narrows the internal divergence between the two modes.

\begin{figure}[t]
   \centering
  \includegraphics[width=\linewidth]{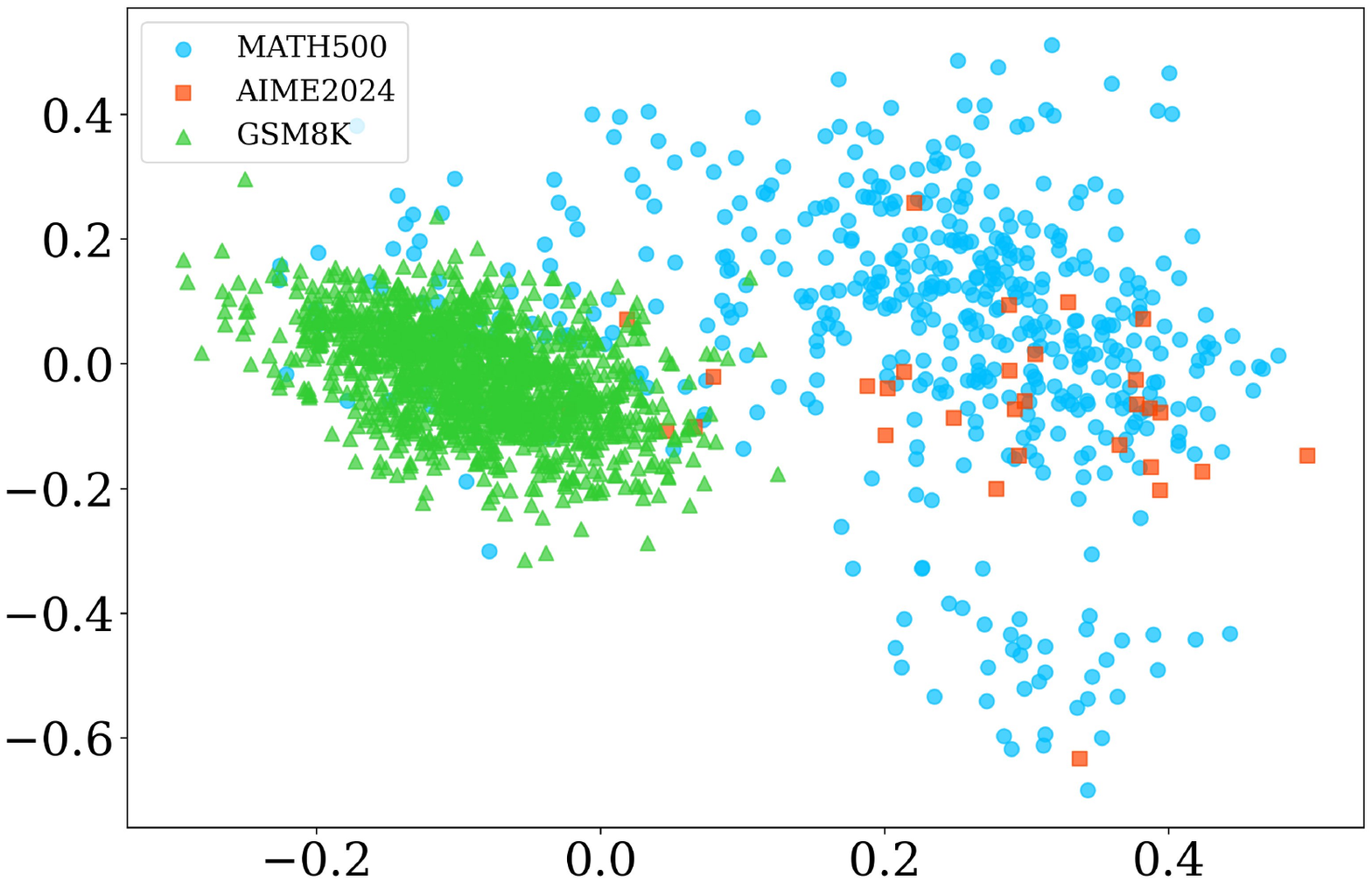}
   \caption{PCA visualization of attention distributions from the 8th layer of \qwen{}.}
   \label{fig:pca_datasets} 
 \end{figure}

To further examine this pattern, we apply PCA to the attention distributions (Figure~\ref{fig:pca_datasets}).
The visualization shows not only a separation between modes but also a coarse clustering by dataset difficulty.
Samples from the simpler GSM8K tend to form a distinct cluster, while samples from the more complex MATH500 and AIME2024 are more dispersed and partially separated.
This suggests that the attention distributions capture features that are related to task difficulty.

\begin{table*}[t] 
 \centering 
 \begin{tabular}{ll ccc} 
 \toprule 

 \multirow{1}{*}{Model} & \multirow{1}{*}{Prompt segment} & \multicolumn{1}{c}{NT} & \multicolumn{1}{c}{ST} & \multicolumn{1}{c}{Difference} \\ 
 \midrule 
 \multirow{2}{*}{\qwq} & Question segment (\%) & 11.88 & 14.82 & 24.75 \\
 & Thought segment (\%) & 66.25 & 59.38 & 10.37 \\ 
 \midrule 
 \multirow{2}{*}{\qwen} & Question segment (\%) & 21.80 & 25.48 & 16.88 \\ 
 & Thought segment (\%) & 40.36 & 38.92 & 3.57 \\ 
 \midrule 
 \multirow{2}{*}{\distill} & Question segment (\%) & 14.42 & 18.84 & 30.65 \\ 
 & Thought segment (\%) & 50.99 & 44.21 & 13.30 \\ 
 \bottomrule 
 \end{tabular} 
 \caption{Average sum of attention on question and thought segments for each LRM, separated by thinking modes. Attention scores are aggregated from layers 21-30.} 
 \label{tab:sectional_attention} 
 \end{table*}

\paragraph{Findings.}
The internal attention states for NT and ST modes become separable in the early layers.
This separation is correlated with task difficulty, but it should not be interpreted as identifying a fine-grained difficulty-control circuit.
The degree of separation varies by model, with QwQ-32B exhibiting the sharpest distinction, while Qwen3-32B shows more blended states.

\subsection{Attentional Focus on Prompt Segments}
\paragraph{Motivation and Methodology.}
Our analysis on DB Index reveals that the internal attention states for NT and ST modes are distinct, 
with divergence emerging from early layers of LRMs. 
This raises the question of what specific features within these attention distributions drive this separation. 
To investigate this, we analyze the LRM's attentional focus by calculating the sum of attention scores originating from the final token of the input.
Specifically, we measure the attention allocated to two key semantic areas: the \textbf{question segment} (containing the original query) and the \textbf{thought segment} (the pre-filled reasoning trigger).

Due to the attention sink phenomenon \cite{attention_sink}, we first exclude the initial tokens (attention sinks) and normalize the scores over the remaining tokens.
To obtain a robust signal, we adapt multi-layer aggregation~\cite{internal_bias} and select a block of middle layers (i.e., layers 21-30) for analysis. 
We choose these layers based on our DB Index analysis (Figure~\ref{fig:db_index}).
They represent the region where the separation between the two modes is both maximal and stable across all LRMs.

Our metric, the total attention a section receives, is calculated from this cleaned and aggregated distribution. 
Let $S'$ be the set of all input token indices excluding the first three. 
The normalized attention sum for a given section is:
\begin{equation}
\label{eq:sectional_attention_final}
\resizebox{.88\hsize}{!}{$
\text{Attn\_Sum}(\text{Section}) = \frac{\sum_{i \in \text{Section} \cap S'} A_{\text{avg}}(x_i)}{\sum_{j \in S'} A_{\text{avg}}(x_j)},
$}
\end{equation}
where $A_{\text{avg}}(x_i)$ is the raw layer-and-head-averaged attention score for token $x_i$ from the final token of the input $t_1$:
\begin{equation}
\label{eq:avg_attention}
\resizebox{.88\hsize}{!}{$
A_{\text{avg}}(x_i) = \frac{1}{N_L} \sum_{l=L_{\text{start}}}^{L_{\text{end}}} \big( \frac{1}{H_l} \sum_{h=1}^{H_l} \text{Attn}_{l,h}(t_1, x_i) \big),
$}
\end{equation}
where $L_{\text{start}}$ and $L_{\text{end}}$ denote the starting and ending layers, respectively.
$N_L$ is the number of aggregated layers.
$H_l$ is the number of attention heads in layer $l$.
$\text{Attn}_{l,h}$ is the attention weight. 
Comparing this metric between NT and ST modes quantifies how LRMs shift attention across prompt segments.

\paragraph{Results and analysis.}
Table~\ref{tab:sectional_attention} presents the results.
We observe a uniform directional shift across all models: entering the ST mode is consistently accompanied by a reallocation of attention resources.
Specifically, the LRM directs significantly more focus to the question segment, with relative increases ranging from 16.88\% (\qwen{}) to 30.65\% (\distill{}).
This increased focus on the problem context comes at the expense of the pre-filled thought segment, which sees a corresponding reduction in attention.

It is worth noting the distinct behavior of \qwen{}. 
Its relative shift is smaller (16.88\% increase on Question, 3.57\% decrease on Thought).
This is due to its high baseline attention: \qwen{} allocates substantial attention (21.80\%) to the question even in the NT mode, suggesting a default tendency to attend to context.
In contrast, \qwq{} and \distill{}, which start with lower question attention, exhibit sharper shifts when the ST mode is triggered.
Despite these variations, the pattern remains consistent: ST cases allocate more attention to the original question and less to the pre-filled thought segment.
We refer to this pattern as retrospective attention allocation.
This term does not imply literal re-reading; rather, it describes an attention-level tendency to retrieve information from the original question before generation.
This suggests that ST behavior uses the problem statement more strongly during explicit reasoning.

\paragraph{Findings.}
This analysis shows that NT and ST modes differ in how attention is allocated across prompt segments.
In ST cases, the model assigns more attention to the original question and less attention to the pre-filled thought segment.
We interpret this as a retrospective attention allocation pattern, which may support explicit reasoning on difficult queries.
This segment-level analysis does not identify fine-grained circuits, but it provides evidence for how the two modes differ internally.

\subsection{Attention Masking for Saving Thinking}

\paragraph{Motivation and Methodology.}
Based on the above findings, we observe a retrospective attention allocation pattern in ST cases, where the model assigns more attention to the question segment.
We hypothesize that this attention pattern is not only correlated with ST behavior, but may also contribute to it.
To test this hypothesis at the segment level, we propose an attention intervention method.
We select samples where the LRM originally exhibited ST behavior and apply a dynamic mask during generation.
Specifically, using PyTorch hooks, we zero out the attention weights for all input tokens belonging to the question segment after the initial prompt processing.
This limits the model's dynamic attention to the question during generation, encouraging it to generate based more on its initial representation and the thought segment.

Specifically, for each generated token $t_k$, we modified the attention weights $\text{Attn}_{l,h}(t_k, x_i)$ from every head $h$ in every layer $l$. 
The weights for all input tokens $x_i$ belonging to the question segment were set to zero, and the remaining weights were re-normalized to sum to one. 
This intervention limits the model's dynamic access to the original question during generation.

\paragraph{Results and Analysis.}
The results in Table~\ref{tab:mask_results} support our hypothesis.
Limiting attention to the question segment suppresses explicit thinking in some ST cases, converting part of them to NT mode.
The intervention is most effective on the simpler \gsmdata{} dataset (e.g., 25.7\% conversion for \distill{}) but less so on the harder \mathdata{}.
Notably, for the extremely challenging \aimedata{} dataset, our intervention failed to induce mode switching.
It suggests that for highly complex problems, the model's internal drive to reason is robust enough to bypass even this attention blockade, which is consistent with the observation that harder samples are less likely to be converted into NT mode.

While we successfully suppressed the thinking behavior in many cases, a critical question remains: does this suppression affect correctness?
Our evaluation on the converted samples shows that forcing the model into NT mode frequently leads to incorrect answers, despite satisfying the length constraint.
This suggests that dynamic attention to the question segment is functionally important for many difficult samples.
We will provide a comprehensive analysis of the trade-off between accuracy and output length across all datasets in Section~\ref{sec:4}.

\begin{table}[t]
\centering
\begin{tabular}{lcc}
\toprule
{Model} & {\gsmdata{}} & {\mathdata{}} \\
\midrule
\qwq{} & 13.5\% & 6.5\% \\
\qwen{} & 16.7\% & 4.8\% \\
\distill{} & 25.7\% & 4.3\% \\
\bottomrule
\end{tabular}
\caption{Percentage of the test samples changed from the ST mode to the NT mode by our method.}
\label{tab:mask_results}
\end{table}

\section{Performance Analysis of LRM Modes}
\label{sec:4}
The ultimate goal of ``save-thinking'' prompts is to improve efficiency without compromising the result. 
However, the analyses in Section~3 suggest that the ST mode is often useful for preserving correctness on complex reasoning tasks.
In this section, we compare the model's performance in its original state (where it naturally enters ST mode) versus the intervention-induced NT state.

\begin{figure*}[!t]
\centering
\includegraphics[width=\textwidth]{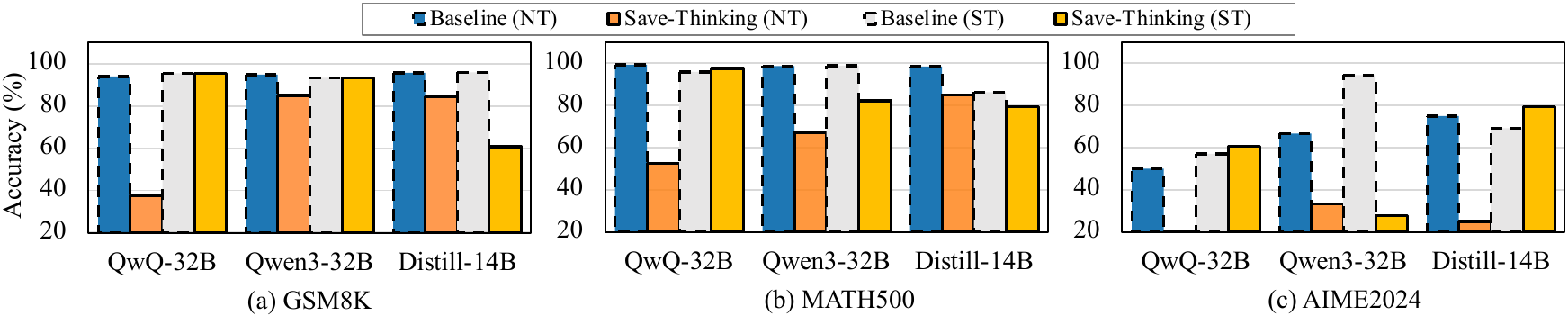}
\caption{Accuracy comparison across three datasets. }
\label{fig:performance}
\end{figure*}

\begin{figure*}[!t]
\centering
\includegraphics[width=\textwidth]{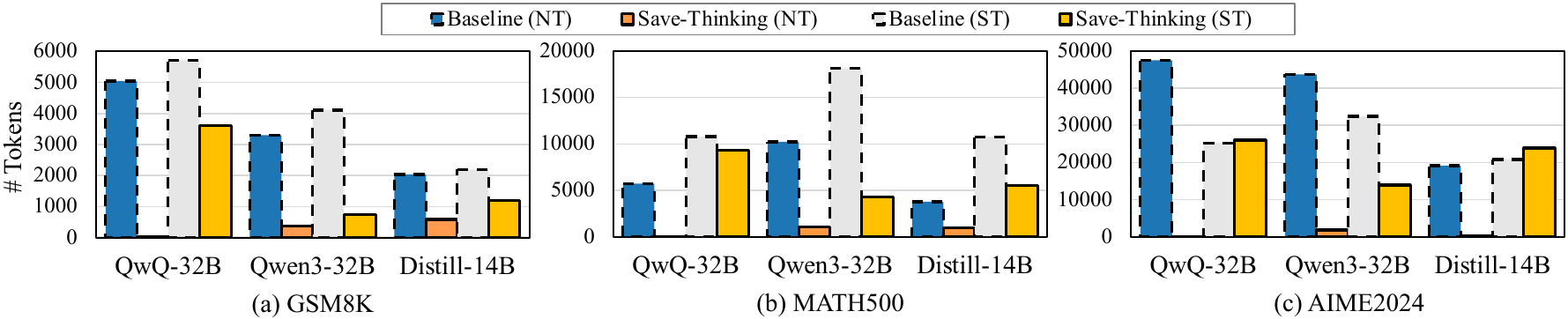}
\caption{The number of output tokens across three datasets. }
\label{fig:performance_len}
\end{figure*}

\subsection{Accuracy and Output Length Analysis}

The results in Figures~\ref{fig:performance} and \ref{fig:performance_len} reveal a clear and consistent trade-off between accuracy and efficiency across all LRMs and datasets.
When LRMs adopt the NT mode, they achieve dramatic length reductions (often $>50\%$) but at a severe cost to accuracy.
For instance, the accuracy of \qwq{} drops by $51.0$  percentage points compared to its baseline, while \distill{} sees a $24.9$ point decrease.
This effect is particularly stark on the most difficult datasets.
For instance, on \aimedata{}, the accuracy of \distill on NT questions falls from a baseline of $75.0\%$ to a  just $25.0\%$. This demonstrates that although the prompt can reduce the output length, performance fails catastrophically when the LRM incorrectly adopts the NT mode for questions requiring reasoning.

In contrast, the ST mode—where the model overrides the instruction—preserves reasoning capabilities.
For RL-trained models (\qwq and \qwen), they maintain accuracy levels comparable to the baseline while still being significantly more efficient than standard generation. 
For instance, on \gsmdata{}, \qwen{}'s ST mode achieves the exact same accuracy as its baseline (93.33\%) but reduces the average answer length by over $82\%$ (from 4,104 to 735 tokens).
This indicates that for RL models, the ``save-thinking'' prompt—when overridden—can successfully act as a conciseness trigger, filtering out verbose reasoning steps while preserving the core logical chain.

However, this benefit does not generalize to the SFT-trained \distill{}.
Unlike the RL models, its ST mode accuracy drops significantly compared to its baseline, and on \gsmdata{} and \mathdata{}, it performs even worse than its own NT mode.
This likely stems from a conflict between the prompt's constraints and the model's training objective. 
Since SFT models are trained to imitate specific ``gold standard'' reasoning trajectories, the external pressure to ``save thinking'' forces the model into an out-of-distribution state that disrupts its learned patterns, leading to degraded performance even when it attempts to think.

In summary, the ``save-thinking'' prompt introduces a fundamental trade-off.
While it offers significant efficiency gains, it also increases the risk of errors on complex tasks.
Models sometimes override the instruction to maintain accuracy, suggesting that explicit reasoning is often important for difficult samples.
Simply shortening the output without reducing the underlying task difficulty can lead to performance degradation.

Manual inspection of NT errors shows that LRMs sometimes produce numerical ``near-misses'', where the answer differs from the correct solution by only a small margin.
(e.g., a single digit). 
This suggests that the NT mode triggers implicit calculation but lacks the necessary self-verification to ensure precision. 
We provide detailed examples in Appendix~\ref{bad_case}.

\subsection{Impact of Attention Masking}
We further evaluate the performance of samples where the NT mode was forcibly induced via our attention masking intervention.
While this successfully constrained the output length to match the natural NT mode, it resulted in a similar drop in accuracy.
For instance, on \gsmdata{}, the accuracy for converted samples was 16/47 for \qwq{}, 3/5 for \qwen{}, and 15/19 for \distill{}.
Performance on the harder \mathdata{} was similarly low: 11/25, 3/4, and 6/11, respectively.
For \qwq{}, limiting dynamic attention to the question segment caused it to fail on the majority of questions (e.g., losing 66\% of cases on GSM8K).
These results suggest that dynamic attention to the question segment is functionally important for many converted samples.
However, this intervention is still coarse-grained and does not isolate the exact heads or circuits responsible for the behavior.

\section{Conclusion}
In this work, we investigated why LRMs sometimes override ``save-thinking'' instructions.
We identify two response modes, NT and ST, and show that ST behavior is associated with task difficulty, higher perplexity at the thinking-termination boundary, and stronger attention to the original question.
Our intervention experiments show that limiting dynamic attention to the question segment can suppress explicit thinking and shorten outputs, but also reduces accuracy on difficult tasks.
These results suggest that explicit reasoning is often useful for correctness, rather than merely redundant verbosity.
They also indicate that save-thinking prompts should not be viewed as a universal shortcut for reducing inference cost.
Instead, controllable LRMs may need adaptive strategies that decide when to shorten reasoning and when to preserve it.
Overall, our findings highlight a tension between instruction following, inference efficiency, and reasoning accuracy, and can inform future work on controllable LRMs and adaptive inference.
\section*{Ethical Considerations}
The datasets and LLMs used in this work are public with permissive licenses.
\section*{Limitations}

We acknowledge three main limitations in this study.
First, this work is primarily a \textbf{mechanistic investigation} into the internal conflict between efficiency instructions and reasoning behaviors. 
While we identify a retrospective attention allocation pattern and the accuracy-efficiency trade-off, we do not propose a novel inference algorithm or a concrete adaptive system to automatically resolve this conflict in production environments.
Second, our evaluation is restricted to \textbf{mathematical reasoning} benchmarks. 
Math problems provide a clear signal for reasoning depth, but other complex domains—such as code generation, creative writing, or agentic planning—may exhibit different attentional dynamics and compliance patterns that are not captured in this analysis.
Finally, our mechanistic analysis operates at the \textbf{layer and segment level}. 
We focused on the macro-distribution of attention (e.g., Question vs. Thought segments) across layers. 
We did not pinpoint specific attention heads or sparse neuronal circuits responsible for the difficulty perception, which limits the immediate applicability of our findings for fine-grained model editing or circuit-level intervention.
Future work can use more fine-grained interventions, such as targeted activation patching or head-level analysis, to further isolate the components responsible for this behavior.

\section*{Acknowledgments}
This work was supported by National Natural Science Foundation of China (No. 62406136) and Natural Science Foundation of Jiangsu Province (No. BK20241246).
\bibliography{custom}

\appendix

\section{Experiment Details}
For all experiments conducted in this study, we employed greedy decoding by setting the do\_sample parameter to False, and the maximum number of tokens allowed for generation was set to 20480. All experiments were performed on a single NVIDIA A800 GPU within a server running Ubuntu 20.04 LTS and equipped with two Intel Xeon Gold 6326 processors and 768 GB of system memory. All models were loaded in bfloat16 precision for all experiments. 

To accelerate inference and obtain the generated outputs efficiently, we utilized the VLLM library. However, since VLLM does not provide access to internal model states such as attention weights during the generation process, we performed an additional forward pass using the original model to retrieve the required internal states after generation was completed. Furthermore, to ensure reproducibility, the random seed was consistently set to 42 across all experiments.

To determine the correctness of the LRM's output, we first extracted the final answer using regular expressions. We then calculated the number of exact matches (EM) with the golden answer. For instances initially judged as incorrect by this automated process, we conducted manual verification to check for any misjudgments by the matching script. Finally, the samples that are exact matches and those manually verified as correct are combined to calculate the final accuracy rate.

\section{Examples of incorrect answers in NT mode}
\label{bad_case}
\begin{table}[!t]
    \centering
    \footnotesize
    \begin{tabular}{p{0.3\linewidth} p{0.35\linewidth} p{0.18\linewidth}}
        \toprule
        \textbf{Questions} & \textbf{Golden answers} & \textbf{Answers} \\
        \midrule

        I have 10 liters of orange drink that are two-thirds water and I wish to add it to 15 liters of pineapple drink that is three-fifths water. But as I pour it, I spill one liter of the orange drink. How much water is in the remaining 24 liters?
        &
        There are 15 x 3/5 = 9 liters of water from the 15 liters pineapple drink. After 1 liter of orange drink was spilled, there were 10 - 1 = 9 liters of orange drink left. Out of the 9 liters, 9 x 2/3 = 6 liters are water. Thus, there are a total of 9 + 6 = 15 liters of water out of the 24 liters. The final answer is $\boxed{15}$.
        &
        The final answer is $\boxed{16}$.
        \\
        \midrule

        A curve is parameterized by $(x,y) = (t^3 + 7, -3t^2 - 6t - 5)$. Find the point the curve passes through at $t = 2$.
        &
        At $t = 2$, $(x,y) = (2^3 + 7, -3 \cdot 2^2 - 6 \cdot 2 - 5) = \boxed{(15,-29)}$.
        &
        The final answer is $\boxed{(15, -23)}$.
        \\
        \bottomrule
    \end{tabular}
    \caption{Examples of incorrect answers from QwQ-32B in the NT mode.}
    \label{tab:error_case_examples}
\end{table}
\begin{table}[h]
    \centering
    \begin{tabularx}{\linewidth}{l X}
        \toprule
        \textbf{Item} & \textbf{Content} \\
        \midrule
        
        \textbf{Question} & 
        There are 9,300 pennies in a cup. What is the total dollar amount in a stack that contains two thirds of the pennies in the cup? \\
        \addlinespace 

        \textbf{Original Output} & 
        Alright, let's tackle this problem step by step... First, I need to figure out how many pennies are in two-thirds of the cup....
        Let me double-check my calculations... is 62. Oh, that's another way to get the same answer... I think that's solid.
        \texttt{</think>}
        
        \underline{The final answer is 62.} \\
        \addlinespace

        \textbf{Intervention Output} & 
        \underline{The final answer is 62.} \\
        
        \bottomrule
    \end{tabularx}
        \caption{A successful example of the attention intervention experiment, presented vertically.}
    \label{tab:intervention_example}
\end{table}
Table~\ref{tab:error_case_examples} presents two examples of incorrect answers generated by the \qwq{} model when it operates in the NT mode. Each row shows the original question, the correct step-by-step solution (Golden answer), and the final answer produced by the model.

As mentioned in the main text, we observe that these incorrect answers are often very close to the correct ones. For instance, in the first case, a multi-step arithmetic problem, the model's output of 16 is off by just a single digit from the correct answer of 15. This suggests an error may occurre in a single final calculation, likely an addition, while the preceding steps were processed with some accuracy. In the second example, when calculating the (x, y) coordinates, the model successfully computes the x-coordinate (15) but fails on the more complex y-coordinate calculation (–23 instead of –29).

This pattern suggests that even without an explicit reasoning chain, the model can perform some correct numerical operations and approximate the structure of the correct solution. However, the lack of a complete thinking chain seems to cause small but critical mistakes in the final steps, leading to a near-miss answer.
\begin{table*}[t]
    \centering

    \begin{tabular}{p{0.22\linewidth} p{0.68\linewidth}}
        \toprule
        \textbf{Component} & \textbf{Description and Criteria} \\
        \midrule
        \multicolumn{2}{p{0.9\linewidth}}{Your task is to analyze the provided mathematical solution to determine if it demonstrates a genuine, deliberative thinking process, rather than just a direct, finalized answer.} \\
        \midrule
        \multicolumn{2}{l}{\textbf{Features of a ``Thinking Process'' (Positive Case, classify as 1)}} \\
        \midrule
        \textbf{Backtracking or Self-Correction} & The text shows an initial approach, recognizes it's incorrect or inefficient, and then explicitly switches to a different method (e.g., ``Wait, that is not right.''). \\
        \addlinespace[0.5em]
        \textbf{Self-Verification} & The text includes explicit checks on its own intermediate steps or the final result to confirm their validity (e.g., ``Let me double-check that calculation,'' or ``Does this answer make sense?''). \\
        \addlinespace[0.5em]
        \textbf{Exploration of Alternatives} & The text considers or mentions multiple possible strategies or solutions before settling on one. It explores the problem space rather than following a single, linear path (e.g., ``Let me try another solution,'' or ``This solution may not good.''). \\
        \addlinespace[0.5em]
        \textbf{Expressions of Uncertainty} & The text uses phrases that indicate ongoing thought, speculation, or a lack of absolute certainty, such as ``I think,'' ``perhaps,'' ``let's assume,'' ``maybe,'' or ``let me try...''. \\
        \addlinespace[0.5em]
        \midrule
        \multicolumn{2}{l}{\textbf{No Thinking Process (Negative Case, classify as 0)}} \\
        \multicolumn{2}{p{0.9\linewidth}}{The solution is a clean, direct, textbook-style set of steps. It is polished and confident, lacking any of the features listed above. It reads like a final, prepared explanation.} \\
        \midrule
        \multicolumn{2}{p{0.9\linewidth}}{\textbf{CRITICAL:} Your entire response must be only the single digit `1` (for thinking process) or `0` (for no thinking process). Do not provide any other text, explanation, or punctuation.} \\
        \bottomrule
    \end{tabular}
        \caption{The detailed prompt and classification criteria for identifying the ST mode, as provided to GPT-4.1.}
    \label{tab:GPT4.1_prompt_details}
\end{table*}

\section{Prompt for Mode Classification}
To obtain consistent NT/ST labels, we used GPT-4.1 as the initial annotator.
The classification was guided by a detailed prompt that specified the criteria for each mode, and the resulting labels were then checked manually.

This prompt, detailed in Table~\ref{tab:GPT4.1_prompt_details}, explicitly outlines the specific criteria and features that define a genuine ST process, such as self-verification or backtracking. Following this automated first pass, we conducted manual verification to review the classifications and ensure the highest level of accuracy. This two-stage approach combines the efficiency of automated annotation with the precision of human oversight, ensuring that the foundational data for our study is both consistent and trustworthy.

\section{Qualitative Examples of Attention Intervention}

This section provides a clear, qualitative example of a successful attention intervention experiment. The goal of this experiment was to test our hypothesis that a LRM's ability to re-attend to the original question is essential for it to enter the ST mode. The following table uses a real example from the \qwq{} model to show how our intervention method works.

Table~\ref{tab:intervention_example} is split into three parts: the original question, the model's initial output, and the output after our intervention was applied.

The ``Original Output'' column shows the model is clearly in ST mode. It doesn't give the answer right away. Instead, it shows its work by breaking the problem down, calculating step-by-step, and even using different methods to double-check its own answer (for example, by saying ``Let me double-check my calculations...''). This long thought process is a perfect example of the ST mode.

An interesting fact about the \qwq model is that it sometimes prints an extra, unneeded \texttt{</think>} tag before giving the final answer. We believe this may happen because its training on the output format was not perfectly complete. However, this unique behavior is actually helpful for our analysis, as it provides another clear signal that helps us distinguish between the ST and NT modes.

\section{Statistical Significance of PPL Differences}
\label{app:ppl-significance}

To further verify the PPL gap between NT and ST cases, we conduct two-tailed independent $t$-tests for each model--dataset pair.
The test compares the PPL values of samples labeled as NT with those labeled as ST under the same model and dataset.
This setting directly matches our analysis in Section~\ref{sec:3}, where the main question is whether ST cases have higher PPL at the thinking-termination boundary.
We focus on GSM8K and MATH500 because both NT and ST groups contain sufficient samples for comparison, while several AIME2024 groups are too small for a reliable test.
For this reason, we exclude AIME2024 from the significance table and treat its PPL pattern as descriptive evidence.

As shown in Table~\ref{tab:ppl-significance}, the differences are statistically significant in five out of six settings.
In particular, all three models show significant NT--ST PPL differences on MATH500, with $p < 0.01$.
On GSM8K, QwQ-32B and Qwen3-32B also show significant differences, while Distill-14B shows a marginal trend.
These results provide additional statistical support for our observation that higher PPL at the thinking-termination boundary is associated with later ST behavior.

\section{AI Assistance Statement}
We used Qwen\footnote{\url{https://qwenlm.github.io/blog/qwen3/}} and Gemini\footnote{\url{https://blog.google/products/gemini/gemini-3-gemini-app/}} for linguistic polishing and stylistic editing of the draft to enhance clarity and readability.
\begin{table}[t]
\centering
\small
\begin{tabular}{llcl}
\toprule
Model & Dataset & $p$-value & Significance level \\
\midrule
QwQ-32B & GSM8K & 0.033 & $p < 0.05$ \\
QwQ-32B & MATH500 & 0.009 & $p < 0.01$ \\
Qwen3-32B & GSM8K & 0.0005 & $p < 0.001$ \\
Qwen3-32B & MATH500 & 0.0065 & $p < 0.01$ \\
Distill-14B & GSM8K & 0.065 & marginal \\
Distill-14B & MATH500 & 0.00002 & $p < 0.001$ \\
\bottomrule
\end{tabular}
\caption{Statistical significance of PPL differences between NT and ST modes.
We conduct two-tailed independent $t$-tests for each model--dataset pair on GSM8K and MATH500.}
\label{tab:ppl-significance}
\end{table}

\end{document}